%% file: acl_latex.tex
\pdfoutput=1

\documentclass[11pt]{article}

\usepackage[]{acl}

\usepackage{times}
\usepackage{latexsym}
\usepackage{authblk}
\usepackage{url}
\usepackage[T1]{fontenc}
\usepackage{siunitx}
\usepackage{latexsym}
\usepackage{times}
\usepackage{float}
\usepackage{footnote}
\usepackage{caption, subcaption}
\usepackage{graphicx}
\usepackage{mathtools}
\usepackage{amsmath}
\usepackage{amssymb}
 \usepackage{booktabs}
 \usepackage{makecell}
 \usepackage{multirow}
\makesavenoteenv{tabular}
\makesavenoteenv{table}
\usepackage{graphicx}
\usepackage{textcomp}
\usepackage{hhline}
\usepackage{colortbl}
\usepackage{latexsym}
\usepackage{standalone}
\usepackage{url}
\usepackage{tikz}
\usepackage{soul}
\usepackage{fixltx2e}
\usepackage{color}
\usetikzlibrary{arrows,decorations.pathmorphing,backgrounds,positioning,fit,petri,calc}
\usepackage{pgfmath}

\definecolor{light-gray}{gray}{0.85}

\DeclareRobustCommand{\best}[1]{{\sethlcolor{light-gray}\hl{#1}}}

\DeclareRobustCommand{\slap4slip}{\textsc{Slap4slip}}

\DeclareMathOperator{\pred}{pred}
\DeclareMathOperator{\total}{total}
\DeclareMathOperator{\reg}{reg}

\usepackage{microtype}

\usepackage[T1]{fontenc}

\usepackage[utf8]{inputenc}

\usepackage{microtype}

\title{Modeling Ideological Salience and Framing in Polarized Online Groups 
with Graph Neural Networks and Structured Sparsity}

\author[*$\ddag$]{Valentin Hofmann}
\author[$\dag$]{Xiaowen Dong}
\author[$\dag$*]{Janet B. Pierrehumbert}
\author[$\ddag$]{Hinrich Sch\"utze}

\affil[*]{Faculty of Linguistics, University of Oxford}
\affil[$\dag$]{Department of Engineering Science, University of Oxford}
\affil[$\ddag$]{Center for Information and Language Processing, LMU Munich \protect\\ \texttt{valentin.hofmann@ling-phil.ox.ac.uk}}

\begin{document}

\maketitle

\begin{abstract}
The increasing polarization 
of online political discourse 
calls for computational tools 
that automatically
detect and monitor ideological divides in
social media. We introduce a minimally supervised method 
that 
leverages
the network structure of online discussion forums, specifically 
Reddit, to detect polarized concepts. 
We model polarization along the dimensions of 
salience and framing, drawing upon insights from moral 
psychology. Our architecture 
combines graph neural networks with structured sparsity learning
and results in representations 
for concepts and subreddits that capture
temporal ideological dynamics such as right-wing and left-wing radicalization.

\end{abstract}

\section{Introduction}

The polarization of online political 
discourse on platforms such as Twitter \citep{Himelboim.2013}, Facebook \citep{Bakshy.2015}, and Reddit \citep{An.2019}
has received increasing attention in the computational social sciences
recently, particularly in the context of Covid-19 \citep{Green.2020}. 
In NLP, a growing body of work has discovered mechanisms 
by which polarization manifests itself linguistically (e.g., \citealp{Demszky.2019}). However, the methods proposed so far rely
on knowing in advance the political orientation of text, a requirement
seldom met in social media.

\def\fascistmainstream{0.2}

\begin{figure}[t!]
        \centering      
        \begin{subfigure}[b]{\fascistmainstream\textwidth}  
          
            \includegraphics[width=\textwidth]{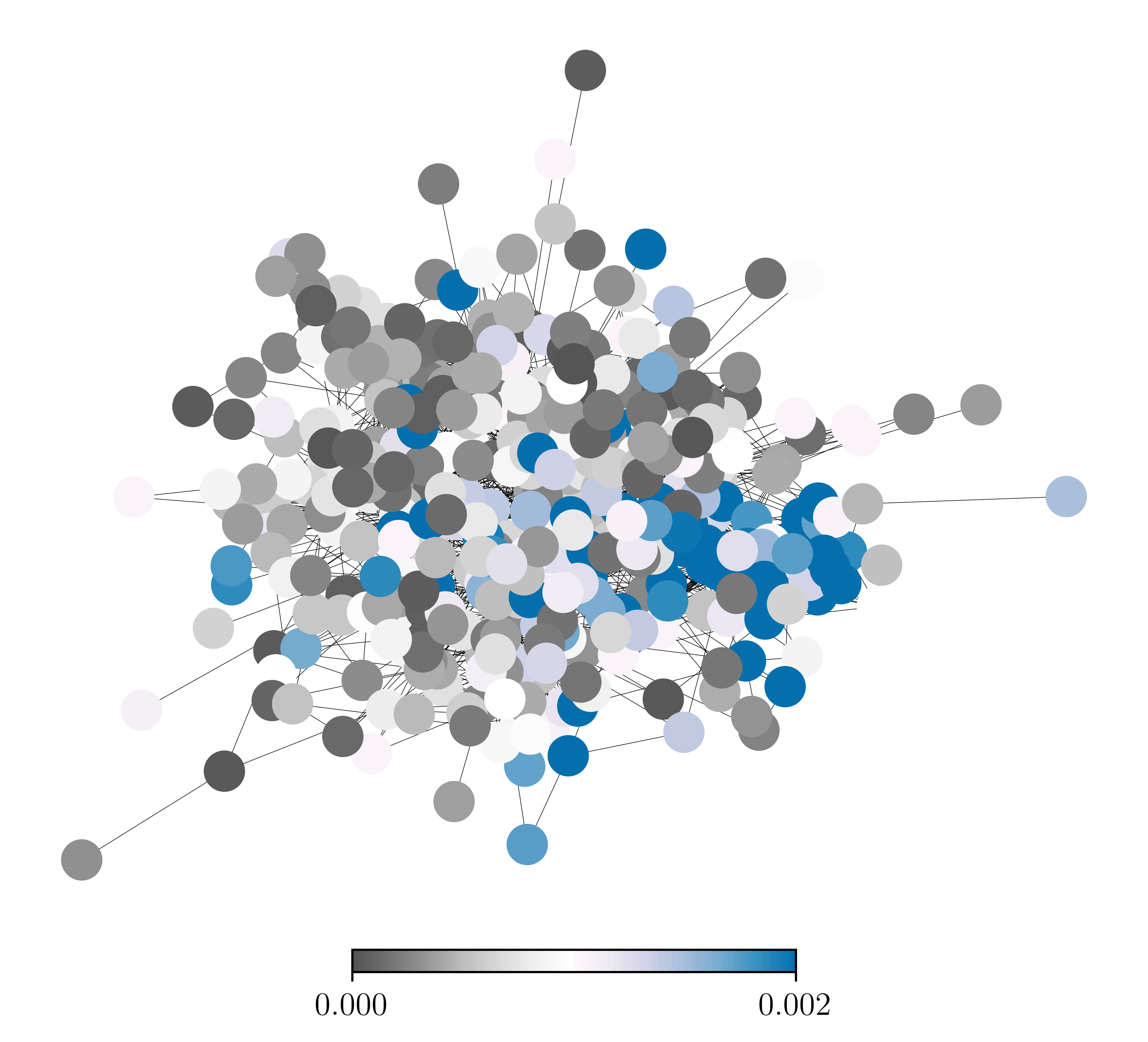}
            \caption[]%
            {{\small \textit{fascist} (salience)}}    
            \label{fig:fascist}
        \end{subfigure}     
        \begin{subfigure}[b]{\fascistmainstream\textwidth}   
         
            \includegraphics[width=\textwidth]{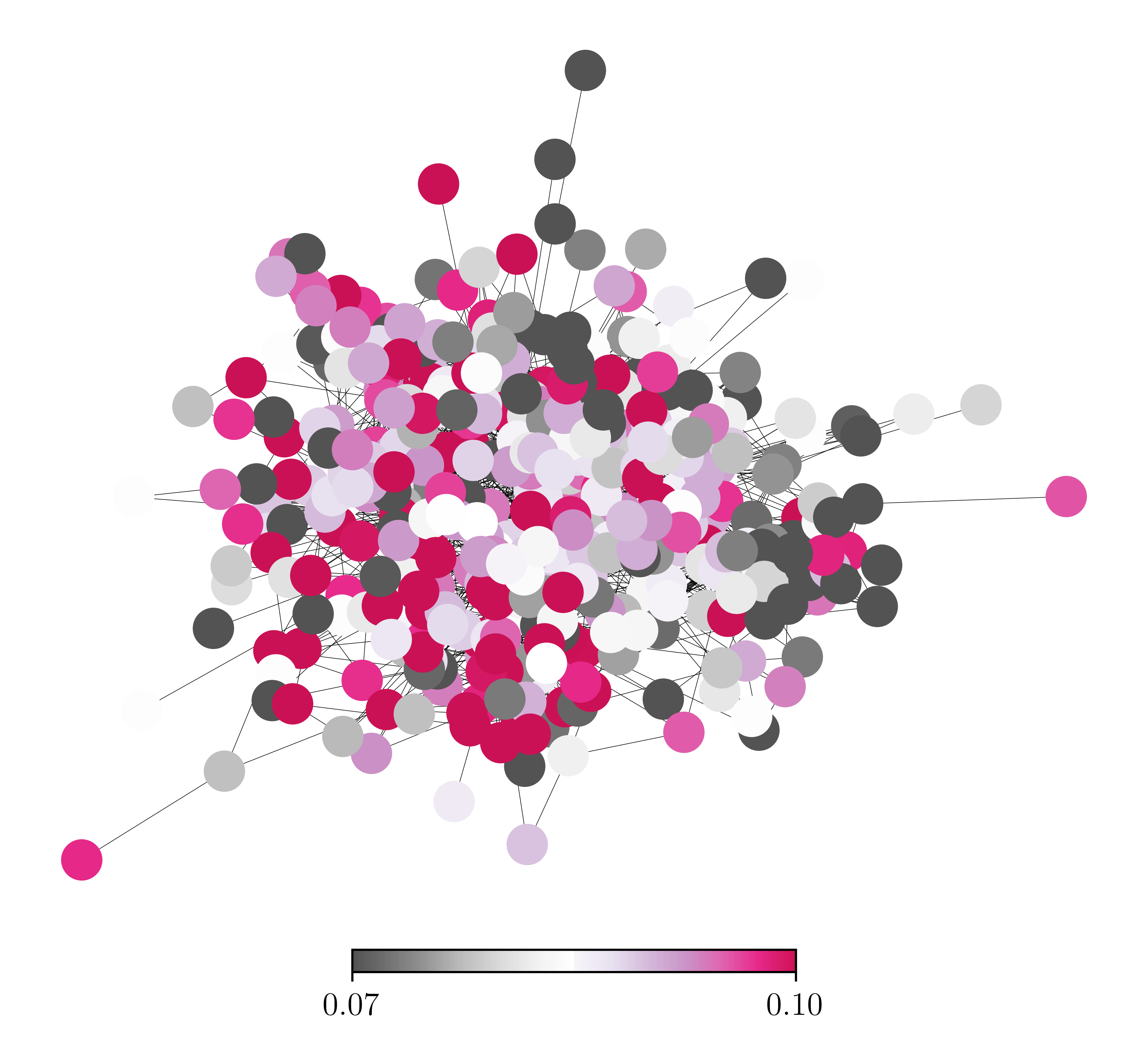}
            \caption[]%
            {{\small \textit{mainstream} (framing)}}    
            \label{fig:mainstream}
        \end{subfigure}
        \caption[]{Examples of concepts polarized along the dimensions of salience (a)
          and framing (b) in Reddit in 2019.
Each circle is a subreddit.
          The values 
        for salience (a) are relative concept frequencies. 
        References to fascism, reflected by higher relative frequencies of \textit{fascist},
        are typical for 
        left-wing
        subreddits (blue region). The values for framing (b)
        are contextualized BERT embeddings projected into
        the moral sanctity/degradation subspace.
         The framing of \textit{mainstream} as degenerate
         is pronounced in 
         right-wing subreddits (magenta region). We can diagnose such patterns using \slap4slip in a minimally supervised way.}
        \label{fig:polarized-networks}
\end{figure}

In this paper, we propose 
\slap4slip (Sparse LAnguage Properties
for Social LInk Prediction), a novel framework
that \textit{fully dispenses with the need for labels}
and instead
leverages
the ubiquitous network structure of online discussion forums to detect polarized concepts,
making it more scalable and lightweight 
than previous methods. 
For example, \slap4slip finds that \textit{fascist} 
and \textit{mainstream} are among the most polarized concepts in Reddit in 2019 (Figure~\ref{fig:polarized-networks}).
We model the polarization of concepts along the dimensions of salience
and framing. For framing, we take into account insights about the 
moral foundations of ideology \citep{Haidt.2004}
and use contextualized BERT embeddings to construct subspaces that capture nuanced biases in 
the way concepts are discussed.

\begin{table*} [t!]\centering
\resizebox{\linewidth}{!}{%
\begin{tabular}{@{}ll@{}}
\toprule
Key term & Explanation\\
\midrule
Polarization & \makecell[{{p{18cm}}}]{By \textit{polarization} we mean the clustering (of nodes, embeddings, etc.) according 
to ideology. Like \citet{Garcia.2015}, we understand it as a non-binary property, i.e., there can be more than two poles. 
A concept
polarized in salience could be a unigram whose relative frequency has two clusters corresponding to liberalism and conservativism.}\\
\midrule
Salience & \makecell[{{p{18cm}}}]{We understand \textit{salience} as the topical prominence with which issues are discussed,
indicating that a substantial importance is (consciously or unconsciously) ascribed to them. Issues that are highly salient
(e.g., for an
online group) tend to be mentioned often, which is reflected by word frequency statistics.} \\ 
\midrule
Framing & \makecell[{{p{18cm}}}]{We use \textit{framing} to refer to the mechanism by which certain aspects of an issue are 
highlighted. If framing patterns are exploited repeatedly (e.g., in an online group), 
this is reflected by word cooccurrence statistics. Due to the 
importance of moral foundations for ideological thinking, this paper focuses on moral framing.}  \\
\bottomrule
\end{tabular}}
\caption{Overview of our key technical terms. See main text
  for more details.}
\label{tab:definition}
\end{table*}

\textbf{Contributions.}
We introduce \slap4slip, a framework to
detect polarized concepts
without information about the political orientation of text.
The specific model we propose for \slap4slip combines 
graph neural networks with 
structured sparsity learning
and identifies in a minimally supervised way (i)
which concepts are the most polarized ones, (ii) whether 
the polarization is due to differences in
salience or framing, and (iii)
which moral foundations are involved (when framing 
is relevant). Drawing on English Reddit data, we evaluate the model \textit{intrinsically} by conducting various experiments and \textit{extrinsically}
by using the found polarized concepts to predict the ideological leaning of US states. The model also learns subreddit embeddings 
that capture temporal dynamics.\footnote{We make our code available at \url{https://github.com/valentinhofmann/slap4slip}.}

\section{Related Work} \label{sec:related-work}

Our study is closely related to previous NLP work on 
\textbf{polarization} \citep{An.2018, Demszky.2019, Shen.2019, 
Roy.2020, Tyagi.2020, Vorakitphan.2020}, but we try
to avoid the need for explicit information about ideologies
(e.g., manual labels)
by leveraging the network structure of online
discussion forums. Besides being more 
readily applicable in practice, this means our method
is not restricted to a small number of opposing
ideologies, making it theoretically more sound \citep{Jackman.2001}.
There is also work
in the computational social sciences
showing that the structure of various types of online social networks  
reflects polarization
\citep{Adamic.2005, Garcia.2015, Garimella.2018}, which has been explained as
a result of homophily, i.e., nodes close to each other 
are likely to share similar views \citep{McPherson.2001}. 
While these studies partition
the network into a small number 
of ideological communities, our method does not require a discretization step.
More broadly, our study 
is related to NLP work on \textbf{ideology} in general \citep{Iyyer.2014, PreotiucPietro.2017, Kulkarni.2018}.

Research in the political sciences has 
discovered \textbf{salience and framing} as two key
dimensions along which the discussion of issues can vary ideologically.
Salience refers to the 
amount of importance attached to an issue by individuals 
\citep{Eulau.1955, Miller.2017}. 
Mass media can impact salience, an effect called 
agenda setting \citep{McCombs.1972}.
Framing refers to the mechanism by which certain 
aspects of an issue are highlighted \citep{Entman.1993, Druckman.2001}. 
Crucially, 
framing is different from sentiment: it reflects what considerations are 
perceived as important, not what stance 
is taken regarding these
considerations \citep{Nelson.1999}.  
Both salience (with a focus on agenda setting) and framing
have been the subject of previous work in NLP \citep{Tsur.2015, Card.2016, Field.2018, Mendelsohn.2021}. 
Here, we use them to characterize differences 
between online groups.

Psychological research has
shown that the fundamental divisions 
between 
different ideologies
are rooted in their views of morality \citep{Lakoff.2008}.
In \textbf{moral foundations
theory} \citep{Haidt.2004, Graham.2011}, this
has been formalized as variation along the moral foundations of 
care/harm, fairness/cheating, loyalty/betrayal, authority/subversion, and sanctity/degradation.
Several studies have shown that moral foundations theory is a suitable
basis for analyzing ideological framing \citep{Johnson.2018, Mokhberian.2020, He.2021}.
We follow
this approach, but as opposed to prior work we operate with contextualized embeddings
that we project into moral embedding subspaces.

Methodologically, we draw on advances in deep learning with 
\textbf{graph neural networks}, specifically graph auto-encoders
\citep{Kipf.2016, Kipf.2017}. In NLP, such graph-based architectures are increasingly
used to include information from social networks for downstream tasks (e.g., \citealp{Mishra.2019, Hofmann.2021c}). Our work differs in that we
combine deep learning on graphs with \textbf{structured sparsity}, a form of regularization 
similar to $\ell_1$ regularization \citep{Tibshirani.1996} that sets entire groups
of parameters to zero \citep{Alvarez.2016}. Structured
sparsity has been used in NLP before \citep{Eisenstein.2011, Murray.2015, Dodge.2019b}, 
but not in connection with graph neural networks.

The precise definition
of the key technical terms in this paper somewhat varies in the literature (e.g., \citealp{Bramson.2016}).
Table \ref{tab:definition} therefore provides a short overview of how we use these terms.

\section{\slap4slip Framework} \label{sec:slap4slip}

The key idea of this paper is to directly leverage the social 
network structure for determining polarized concepts.\footnote{We define concepts as topics, issues, and public figures discussed in online groups.} 
We introduce a novel framework called \slap4slip
(Sparse LAnguage Properties
for Social LInk Prediction)
whose goal it is to model the structure of social networks 
in a data-driven way that obviates the need for extensive
human annotation or partitioning the network into communities.
  \slap4slip is a general framework to detect the most
salient types of linguistic variablity in social networks and 
is in principle applicable in any scenario involving 
social networks with textual data attached to each node. In this paper, we show that for polarized online discussion forums, 
\slap4slip can be used to find polarized concepts.

Let $\mathcal{G} = (\mathcal{V}, \mathcal{E})$ be a network consisting
of a set of nodes $\mathcal{V}$ representing social entities and a set 
of edges $\mathcal{E}$ representing connections between the
social entities.
We denote with $\mathbf{A} \in \mathbb{R}^{|\mathcal{V}| \times |\mathcal{V}|}$ the adjacency matrix of $\mathcal{G}$. 
Let $\mathcal{C}$ be a set of 
word $n$-grams denoting 
concepts (e.g., political issues like \textit{gun control}). Here, we confine ourselves to subreddits 
for $\mathcal{V}$ and unigrams and bigrams
for $\mathcal{C}$, but \slap4slip
is applicable in other scenarios (e.g., for networks of people
or concepts extracted from text in a more complex manner). We define a function $\psi_l: \mathcal{V} \times \mathcal{C} \rightarrow \mathbb{R}$ 
that assigns to each node $v_i \in \mathcal{V}$ and
concept $c_j \in \mathcal{C}$ the value 
of a linguistic property $l$ observed for $c_j$ in $v_i$. $\psi_l$ can be represented
as a matrix in $\mathbb{R}^{|\mathcal{V}| \times |\mathcal{C}|}$,
\begin{equation*}
\boldsymbol{\Psi}_l = \begin{bmatrix} 
    \psi_l(v_1, c_1) & \dots & \psi_l(v_1, c_{|\mathcal{C}|}) \\
    \vdots & \ddots &  \vdots\\
   \psi_l(v_{|\mathcal{V}|}, c_1) &  \dots      & \psi_l(v_{|\mathcal{V}|}, c_{|\mathcal{C}|})
    \end{bmatrix},
\end{equation*}
where each column is a graph signal \citep{Dong.2020b} over $\mathcal{G}$ determined by $c_j$ and $\psi_l$. For example, if we chose $l$ to be the frequency count, $\psi_l$ 
would indicate how often each
concept occurred in the text attached to each node of the network.

\begin{figure}
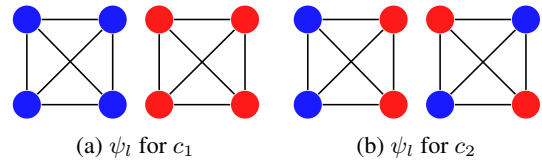

        \centering
        \begin{subfigure}[b]{0.225\textwidth}
            \centering
            \includestandalone[width=0.9\textwidth]{tutorial_1}%
  \caption[]%
            {{\small $\psi_l$ for $c_1$}}   
  \label{fig:tutorial-1}
        \end{subfigure}
        \begin{subfigure}[b]{0.225\textwidth}  
            \centering 
            \includestandalone[width=0.9\textwidth]{tutorial_2}
            \caption[]%
            {{\small $\psi_l$ for $c_2$}}    
            \label{fig:tutorial-2}
        \end{subfigure}
      
        \caption[]{Example for the prediction of graph structure from a
linguistic property. The figures show
          $\psi_l$ for concepts $c_1$ and $c_2$ 
        on a toy graph, with $l$ chosen 
        to be the frequency count represented by node color (identical colors 
        mean identical frequencies). The edges can be fully predicted from $\psi_l$
        for $c_1$ but not $c_2$.}
        \label{fig:tutorial}
\end{figure}

The goal of \slap4slip
is to find the subset of concepts $\mathcal{C}^* \subseteq \mathcal{C}$ that 
best meets the following two 
desiderata: 
(i) given a linguistic property $l$, the signals imposed on $\mathcal{G}$ by $\psi_l$ and the concepts in $\mathcal{C}^*$ should allow for optimal predictions about the structure of $\mathcal{G}$, specifically $\mathcal{E}$; (ii) the number of concepts in $\mathcal{C}^*$ should be minimal.\footnote{Desideratum (i) is conceptually similar 
to measures of opinion polarization 
on networks \citep{Matakos.2017}.} In practice, we treat this as a constrained optimization problem \citep{Bertsekas.1982}, i.e., we use (i) as 
the objective and impose (ii) as a hard constraint on $|\mathcal{C}^*|$.

As an
example, consider the network in Figure
\ref{fig:tutorial}. The network consists of two connected components
of four edges each,
with no edges between the components. 
$\mathcal{C}$ consists of the two concepts $c_1$ and $c_2$. Taking the frequency count
as linguistic property $l$ and displaying it
with the color of nodes, $\psi_l$ results in the two signals shown in Figure \ref{fig:tutorial}. We can see
that the signal of 
concept $c_1$ alone allows for a perfect prediction
of the network structure according to the decision rule
\begin{equation*}
\mathbf{A}_{ij} = \begin{cases}
 1 & \text{if } \psi_l(v_i, c_1) = \psi_l(v_j, c_1) \\    
0 & \text{otherwise}.
\end{cases}
\end{equation*}
Since $c_2$ cannot achieve a perfect prediction, $\mathcal{C}^* = \{c_1\}$ is the 
optimal solution. Notice the variance of $\psi_l(v_i, c_j)$ is identical
for both concepts and does not represent
a good distinguishing factor. Notice also that 
the optimal solution is not necessarily unique: there might be another concept $c_3$ 
with a similar frequency count distribution as $c_1$ such that $\mathcal{C}^* = \{c_3\}$ would
also be an optimal solution.

\section{Model} \label{sec:models}
 
We draw upon Reddit Politosphere \citep{Hofmann.2022}, 
a pseudonymized dataset based on Reddit covering 
605 political subreddits (e.g., \texttt{politics})
from 2008 to 2019.\footnote{\url{https://zenodo.org/record/5851729} (CC-BY 4.0 license)}
For each year, Reddit Politosphere 
contains (i) all comments made to the subreddits and 
(ii) an unweighted graph with the subreddits as nodes and
edges 
computed by applying statistical backboning to the counts of users shared between subreddits.
Subreddits that have 
disproportionately many users in common are likely to be ideologically similar \citep{Kumar.2018}.
To ensure robust training, we only use years in which the graph has at least 100 
nodes (2013 to 2019). See Appendix~\ref{app:stats} for summary statistics. The high 
modularity values indicate that the graphs are polarized \citep{Kirkland.2013}.

We propose a neural architecture 
that uses information about concept-level salience and framing
to predict links between subreddits
while reducing the number of considered concepts as far as possible.
Since the links reflect ideological similarity,
this should result in a compact
set of concepts that is maximally informative about
ideology. The performance on link 
prediction makes it straightforward to compare the quality of different models.

\textbf{Determining
  concepts.} To obtain the
concepts $\mathcal{C}$, we create for 
each year unigram and bigram vocabularies 
of political comments taken from Reddit Politosphere and non-political comments sampled in equal size 
from the default 
subreddits.\footnote{A set of topically diverse subreddits (e.g., \texttt{Fitness})
users used to be subscribed to automatically. We remove \texttt{news} and \texttt{worldnews} since they also contain political content. 
We retrieve the default subreddits from the Pushshift Reddit Dataset \citep{Baumgartner.2020}.} 
To eliminate unigrams and bigrams typical of discussions but not
relevant to salience and framing (e.g., \textit{dont think}),
we only consider unigrams and bigrams that appear more often within than outside of noun phrases as detected
by a noun phrase chunker \citep{Honnibal.2020}.
Based on their frequencies within the political and non-political comments, we compute mutual information scores
for all unigrams and bigrams and take the top 1,000 unigrams and bigrams for $\mathcal{C}$.
This and all other steps are done separately for each year, 
i.e., we extract year-wise
concepts and train year-wise models.

\textbf{Modeling salience and framing.} The first part of the architecture models 
$\psi_l$, i.e., it extracts linguistic 
information related to salience and framing
from the subreddits and maps them to scalar
representations. In the resulting matrix $\boldsymbol{\Psi}_l$, each column is
a signal on the entire graph defined by one concept, and each row 
is a vector for one subreddit defined by all concepts in $\mathcal{C}$ (Section~\ref{sec:slap4slip}).

To model ideological salience, we measure the relative frequency of concepts 
\begin{equation*}
s(v_i, c_j) = \frac{n(v_i, c_j)}{\sum_k n(v_i, c_k)},
\end{equation*}
where $n(v_i, c_j)$ is the frequency count
of concept $c_j$ in subreddit $v_i$. Variations in the relative frequency of a concept 
that are strongly correlated
with the social network structure indicate that the concept is
used with systematically higher frequency in certain regions of the social network, 
potentially caused by its elevated place 
within the ideologies of the subreddits in question.

To model ideologically-driven framing, we use BERT (base, uncased; \citealp{Devlin.2019}) 
and obtain average contextualized embeddings $\mathbf{e}(v_i, c_j)$ 
for each subreddit $v_i$ and concept $c_j$. 
Furthermore, we use the Moral Foundations Dictionary \citep{Frimer.2017}
and obtain for each moral foundation $m_k$ (e.g., authority/subversion) average 
contextualized embeddings for the 10 highest-ranked words of both poles.\footnote{\url{https://osf.io/ezn37} (CC-BY 4.0 license)} 
Similar to \citet{Bolukbasi.2016}, we 
perform PCA on the 20 average contextualized embeddings for each 
$m_k$ and use the first principal component as the subspace representation $\mathbf{e}(m_k)$. 
This allows us to project the subreddit-specific average contextualized concept embeddings
$\mathbf{e}(v_i, c_j)$
into the five moral subspaces,
\begin{equation*}
p_k(v_i, c_j) = \cos \left(\mathbf{e}( v_i, c_j ), \mathbf{e} ( m_k )  \right).
\end{equation*}
$p_k(v_i, c_j)$ reflects how relevant the moral foundation $m_k$ is for the contexts 
in which concept $c_j$ occurs in subreddit $v_i$ (see Appendix~\ref{app:subspaces} for 
 further details and a systematic evaluation). The moral foundations 
are expected to be relevant for the framing of concepts to differing degrees. We therefore
compute concept-specific weighted sums,
\begin{equation*}
f(v_i, c_j)  = \sum_k \pi_k^{(c_j)} p_k(v_i, c_j),
\end{equation*}
where $\sum_k \pi_k^{(c_j)} = 1$ and $\pi_k^{(c_j)} \geq
0$. $f(v_i, c_j)$ is an aggregate indicator of how important
moral framing is for concept $c_j$ in $v_i$.
The parameters
$\pi_k^{(c_j)}$ are optimized during
training.

Salience and framing can be of different importance for different concepts, i.e.,
there might be concepts with identical values of $s(v_i, c_j)$ across all subreddits 
but maximally polarized values of $f(v_i, c_j)$ (or vice versa).
To capture this,
we combine $s(v_i, c_j)$ and $f(v_i, c_j)$ in a weighted sum,
\begin{equation*}
o(v_i, c_j) = \alpha^{(c_j)} s(v_i, c_j) + (1 - \alpha^{(c_j)})  f(v_i, c_j),
\end{equation*}
where $0 \leq  \alpha^{(c_j)} \leq 1$ is again a concept-specific parameter that is
optimized during 
training. $o(v_i, c_j)$ indicates the overall activation 
of concept $c_j$ in $v_i$ (i.e., both due to salience and framing).
Two important points must be stressed. First, $\pi_k^{(c_j)}$ and $\alpha^{(c_j)}$ are specific for concepts but identical for subreddits: e.g., if a concept $c_j$ has $\alpha^{(c_j)} = 1$, this means that only information from $s(v_i, c_j)$ is used for all subreddits. Second, values for $o(v_i, c_j)$ are comparable across subreddits but not across concepts: since $\pi_k^{(c_j)}$ and $\alpha^{(c_j)}$ differ between concepts, differences in 
$o(v_i, c_j)$ are not meaningful for different concepts (see Section~\ref{sec:experiments} for 
examples).
To get the final concept representation that is passed to 
subsequent parts of the model, we set $\psi_l = o$, i.e., each entry in $\boldsymbol{\Psi}_l$
contains the value of $o(v_i, c_j)$ for subreddit $v_i$ and concept $c_j$.

\textbf{Graph neural network.} To predict the links in 
$\mathcal{G}$, we use a graph neural network \citep{Wu.2021}, specifically a graph auto-encoder \citep{Kipf.2016}, which 
takes as input the matrix $\boldsymbol{\Psi}_l$
as well as $\mathcal{G}$'s adjacency matrix $\mathbf{A}$.

The encoder consists of a two-layer graph convolutional network \citep{Kipf.2017}. 
In each layer, the subreddit representations $\mathbf{H}^{(d)}$
are updated according to the propagation rule
\begin{equation*}
\mathbf{H}^{(d+1)} = \sigma \left( \tilde{\mathbf{D}}^{-\frac{1}{2}} \tilde{\mathbf{A}}
\tilde{\mathbf{D}}^{-\frac{1}{2}} \mathbf{H}^{(d)} \mathbf{W}^{(d)}
  \right),
\end{equation*}
where $\tilde{\mathbf{A}} =  \mathbf{A} + \mathbf{I}$ is $\mathcal{G}$'s adjacency matrix with added
self-loops, $\tilde{\mathbf{D}}$ is the degree matrix of $\tilde{\mathbf{A}}$,
and $\mathbf{W}^{(d)}$ is the weight matrix of layer $d$. $\sigma$ is the activation function, 
for which we use a rectified linear unit \citep{Nair.2010} after the first
and a linear activation (no non-linearity) after the second layer. 
We set $\mathbf{H}^{(0)} = \boldsymbol{\Psi}_l$. In our architecture, 
$\mathbf{Z} = \mathbf{H}^{(2)}$ is the output of the encoder.
Graph convolutions are mathematically equivalent to Laplacian 
smoothing \citep{Li.2018}, which is an important property for our architecture:
if a concept does not occur in a subreddit, it 
ensures that the subreddit receives a high-quality representation by drawing on
the neighboring subreddits.

In the decoder, we compute the reconstructed adjacency matrix, $\hat{\mathbf{A}}$,
according to
\begin{equation*}
\hat{\mathbf{A}} = \sigma \left( \mathbf{Z}  \mathbf{Z}^\top \right),
\end{equation*} 
where we use the sigmoid for $\sigma$. $\hat{\mathbf{A}}$ is then used to compute 
a prediction loss, $\mathcal{L}^{(\pred)}$.

\textbf{Structured sparsity.} Following the \slap4slip framework,
we want to reduce the number of concepts in $\mathcal{C}$. In the described architecture, 
this amounts to reducing the number of columns in $\boldsymbol{\Psi}_l$.
We want to achieve this as part of training, 
using structured sparsity learning, specifically group lasso regularization \citep{Yuan.2006}, to set entire rows of the weight matrix $\mathbf{W}^{(0)}$ to zero. Writing $\mathbf{W}^{(0)} 
= [  \mathbf{w}^{(0)}_1, \dots, 
\mathbf{w}^{(0)}_{|\mathcal{C}|} ]^\top $ as a series of row vectors, 
we define the regularization penalty as
\begin{equation*}
\mathcal{L}^{(\reg)} = \sum_{j=1}^{|\mathcal{C}|} \| \mathbf{w}^{(0)}_j \|_2.
\end{equation*}
This is a mixed $\ell_1$/$\ell_2$ regularization (the $\ell_1$ norm of the
row $\ell_2$ norms) that leads to sparsity on the level of rows. When all 
entries in a row $\mathbf{w}^{(0)}_j$ are zero, this has the effect of
removing concept $c_j$ from $\mathcal{C}$. We compute the final loss as
\begin{equation*}
\mathcal{L}^{(\total)} = \mathcal{L}^{(\pred)} + \lambda \mathcal{L}^{(\reg)}, 
\end{equation*}
where $\lambda > 0$ is a hyperparameter
controlling the intensity of the $\ell_1$/$\ell_2$ regularization.

\section{Experiments} \label{sec:experiments}

\textbf{Setup.} For each year, we split $\mathcal{E}$ into 60\% train, 
20\% dev, and 20\% test edges. We always use the train edges for the 
adjacency matrix $\mathbf{A}$ that is passed to the model, i.e., only the to-be-predicted edges differ between 
train, dev, and test.
 For dev and test, we randomly 
sample non-edges $(v_i, v_j)
\not \in \mathcal{E}$ as negative examples such that edges and non-edges are balanced in both sets (50\% positive, 50\% negative).
For training,
we 
sample non-edges in every epoch (i.e., the set of sampled non-edges changes in every epoch). During test, we rank all edges according to their predicted scores. See Appendix \ref{app:hyperparams}
for hyperparameter details.

In this paper, we use sparsity as a hard constraint on the number
of concepts with non-zero row weights in $\mathbf{W}^{(0)}$, 
i.e., we only consider models for which $|\mathcal{C}| \leq \theta_{|\mathcal{C}|}$, where 
$\theta_{|\mathcal{C}|}$ is the sparsity threshold.
We initially set $\theta_{|\mathcal{C}|} = 150$ but later analyze
its impact in greater detail.

The model is trained with binary cross-entropy as $\mathcal{L}^{(\pred)}$ and Adam \citep{Kingma.2015} 
as the optimizer. 
Since 
$ \mathcal{L}^{(\reg)}$ is non-differentiable, we use
proximal gradient descent \citep{Parikh.2013}.
 We
approximate the 
weighted proximal operator of the $\ell_1$/$\ell_2$ norm
using the Newton-Raphson algorithm \citep{Deleu.2021}.
We use area
under the curve (AUC) for model evaluation. We refer to our model as \textbf{SF-SGAE} (Salience/Framing
Sparse Graph Auto-Encoder).

\textbf{Intrinsic evaluation.} We compare SF-SGAE against three ablated models: one where we use only salience, i.e., $\psi_l = s$ (S-SGAE), 
one where we use only framing, i.e., $\psi_l = f$ (F-SGAE), and
one where we use both types of information 
but replace the graph convolutions with linear layers (SF-SLAE). Furthermore, we implement a model that is identical to SF-SGAE but does
not use sparsity, i.e., $|\mathcal{C}|$ is not reduced (SF-GAE).

\begin{table} [t!]\centering
\resizebox{\linewidth}{!}{%
\begin{tabular}{@{}lrrrrrrrr@{}}
\toprule
Model & 2013 & 2014 & 2015 & 2016 & 2017 & 2018 & 2019 & $\mu\pm\sigma$\\
\midrule
SF-SGAE  & \best{.890} & \best{.895} & \best{.895} & \best{.923} & \best{.937} & \best{.908} & \best{.934} & .912$\pm$.018\\ 
\midrule
S-SGAE & .886 & .890 & .853 & .875 & .894 & .864 & .925 & .884$\pm$.022\\ 
F-SGAE & .875 & .893 & .878 & .885 & .905 & .875 & .917 & .890$\pm$.015\\ 
SF-SLAE & .653 & .810 & .754 & .781 & .764 & .729 & .752 & .749$\pm$.046\\ 
\midrule
SF-GAE  & .829 & .797 & .871 & .916 & .898 & .866 & .933 & .873$\pm$.044\\ 
\bottomrule
\end{tabular}}
\caption{Test performance (AUC). SF-SGAE outperforms S-SGAE, F-SGAE, and SF-SLAE. 
It performs similarly to or better than SF-GAE despite using only a fraction of concepts. 
Best score per column in gray. See Appendix \ref{app:dev} for dev performance.}  \label{tab:auc}
\end{table}

SF-SGAE clearly---and substantially on some years---outperforms the ablated models
(Table \ref{tab:auc}). This shows that jointly modeling salience
and framing captures polarization better than
only modeling one of the two. Between S-SGAE and F-SGAE, there is 
no clear winner, although F-SGAE performs slightly better overall. SF-SLAE performs
substantially worse than all other models, which indicates
that the Laplacian smoothing in the form of graph convolutions
is a crucial component of the model.
SF-SGAE also outperforms SF-GAE on test,
suggesting that $\mathcal{C}^*$ allows
for a more robust generalization  
than the larger but noisier $\mathcal{C}$.

\begin{figure}[t!]
        \centering
        \includegraphics[width=0.4\textwidth]{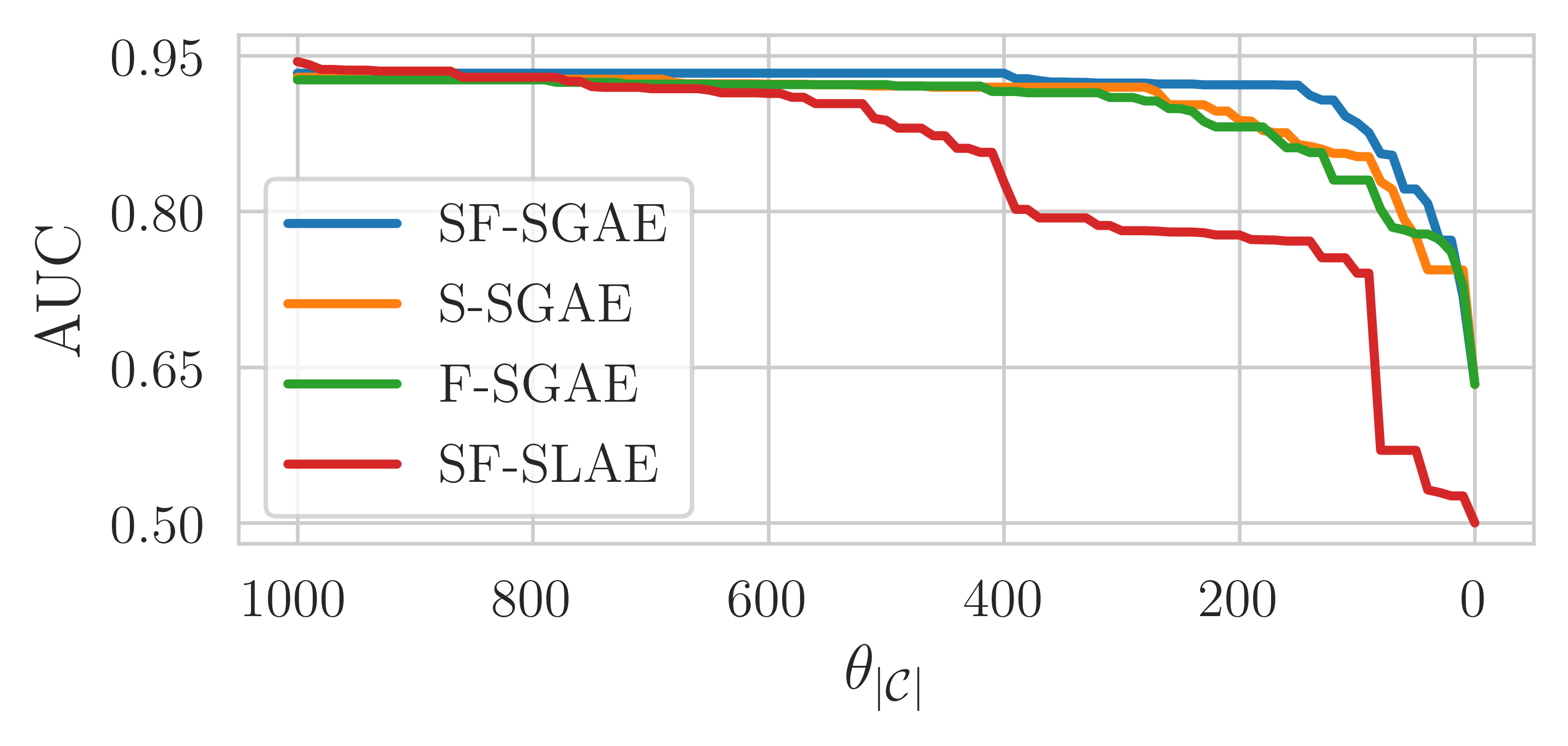}      
        \caption[]{Impact of  sparsity threshold
          $\theta_{|\mathcal{C}|}$ on performance (AUC) on dev for 2016. 
          SF-SGAE performs better than any other model in the sparse regime ($\theta_{|\mathcal{C}|} \leq 200$), 
        showing that it better captures polarization. Plots for all years are provided in Appendix \ref{app:sparsity}.}
        \label{fig:threshold}
\end{figure}

How does the sparsity threshold $\theta_{|\mathcal{C}|}$ impact model performance? The answer to 
this question indicates how many concepts
are required to capture the central ideological divides in the data. We vary $0 \leq \theta_{|\mathcal{C}|} \leq 1000$
and measure the performance (AUC) of the four sparsifying models on dev (Figure \ref{fig:threshold}).
First, we find that for the models
using graph convolutions, reducing $|\mathcal{C}|$ to approximately 200 
concepts does not hurt performance. 
For the model without  graph convolution, on the other hand, performance
starts to drop already around 400 concepts. This makes intuitive sense:
given that the graph convolutions act as a form of smoothing, 
less concepts are needed for a reliable feature vector for each subreddit.
Second, the advantage of SF-SGAE lies not only in its higher performance
in the sparse regime but also in its ability to reduce $|\mathcal{C}|$ much further than any of the other models 
given a performance threshold. This again demonstrates that 
a joint model of salience and framing results in richer information, making it possible to reduce the number of concepts further.

\textbf{Extrinsic evaluation.} The fact that \slap4slip is a minimally supervised framework   
makes it challenging to evaluate the correctness of our model.
While the performance on link prediction indicates 
how well $\mathcal{C}^*$ captures the polarized structure of the social network, 
it is not a direct measure of ideological polarization. There is also 
no ground-truth dataset against which $\mathcal{C}^*$ could be compared. We therefore devise
 an alternative extrinsic evaluation method. 
 Specifically, we use DW-NOMINATE \citep{Poole.1985, Poole.1997}, a quantitative measure of the ideological polarization of members of the US Congress 
 based on their roll-call voting behavior. Recently, a large dataset of DW-NOMINATE scores
 has been made publicly available \citep{Lewis.2021}.

We first create a dataset 
with all comments from subreddits dedicated to US state-level politics (e.g., \texttt{TexasPolitics}) in 2018.\footnote{We choose the larger subreddit in the case of multiple state-level subreddits. We retrieve the subreddits from the Pushshift Reddit Dataset \citep{Baumgartner.2020}.} 
We discard subreddits with less than 250 comments, resulting in a set of 28 subreddits. 
For each state, we then compute the average DW-NOMINATE score of its representatives in the lower house 
of the 116th US Congress (elected in November 2018). The average DW-NOMINATE is 
a continuous measure of the ideological leaning of a state and ranges between \textminus 0.399 for Massachusetts (very liberal)
and 0.467 for Idaho (very conservative). Notice that 
this score reflects the state-level voting shares to a certain extent (since it is averaged over the representatives elected by a state) while at the same time being more fine-grained (since representatives of the same party can differ ideologically). Finally, for each state-level subreddit $v_i$, we 
extract $s(v_i, c_j)$ for (i) the $d$ concepts $c_j$ from $\mathcal{C}^*$ with the highest 
frequency across all state-level subreddits and (ii) $d$ frequency-matched concepts $c_j$ sampled from $\mathcal{C} \setminus \mathcal{C}^*$.\footnote{For $\mathcal{C}^*$, we only consider concepts for which $\alpha^{(c_j)}= 1$, i.e., the polarization is captured by $s(v_i, c_j)$ alone.} We set $d = 5$.\footnote{Results are robust with respect to the exact selection of $d$.}
If the concepts from $\mathcal{C}^*$ are better predictors of the average DW-NOMINATE scores than the concepts from $\mathcal{C} \setminus \mathcal{C}^*$, this indicates
that the model has learned a correct split into more versus less  
polarized concepts.

To test this empirically, we compute the absolute value of Pearson's $r$ between $s(v_i, c_j)$ and the DW-NOMINATE scores. We find a higher correlation for the concepts from $\mathcal{C}^*$ ($\mu = 0.285$, $\sigma = 0.062$) than for the 
concepts from $\mathcal{C} \setminus \mathcal{C}^*$ ($\mu = 0.126$, $\sigma = 0.121$), a difference that is shown to be significant ($p < 0.05$) by a two-tailed $t$-test. This indicates that the concepts in $\mathcal{C}^*$ reflect the polarization of US politics
better than the concepts in $\mathcal{C} \setminus \mathcal{C}^*$.

\begin{figure}[t!]
        \centering
        \includegraphics[width=0.4\textwidth]{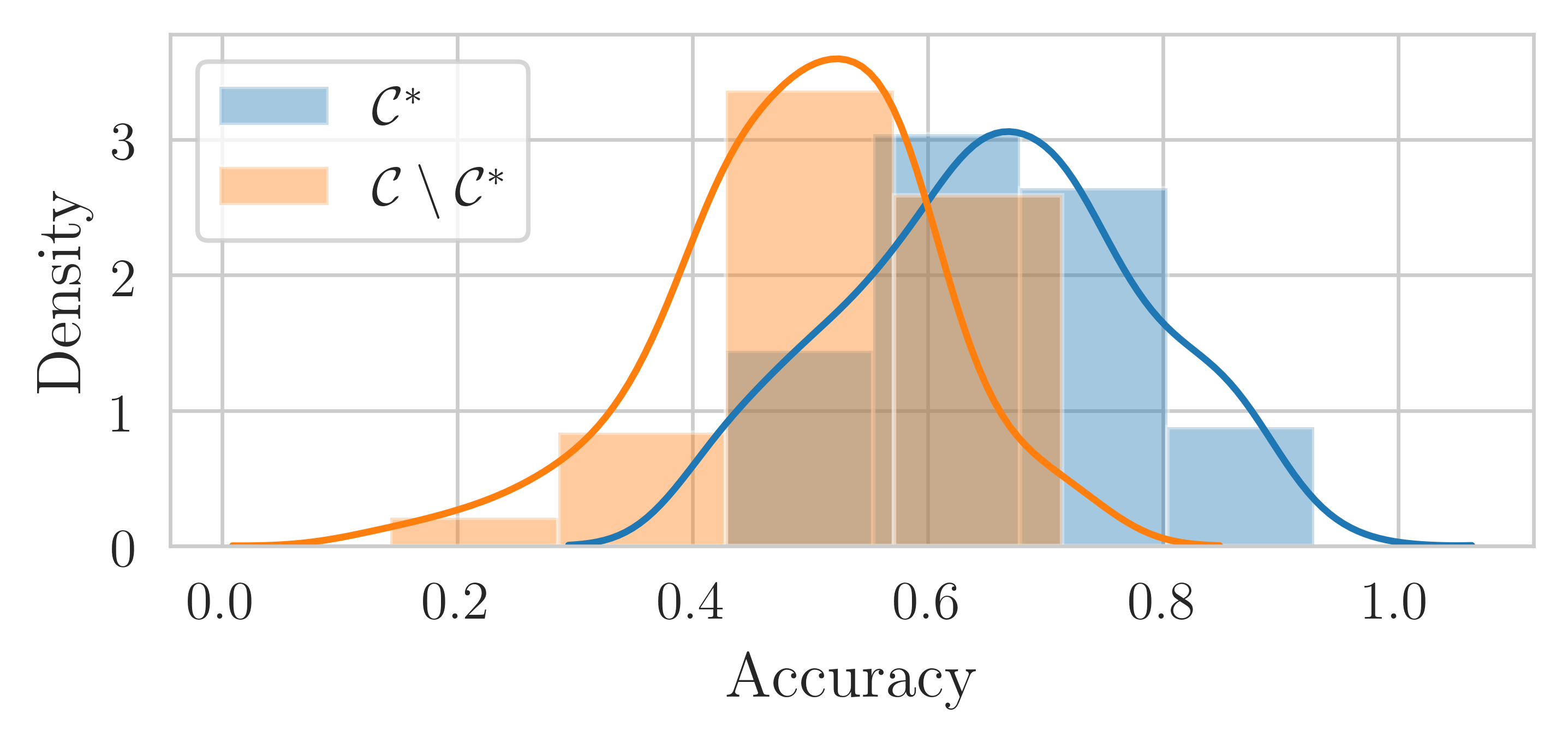}     
        \caption[]{Performance on ideology prediction. The figure shows 
        the distribution of accuracies for 100 models trained with relative frequencies 
        of the concepts from $\mathcal{C}^*$ versus the concepts from $\mathcal{C} \setminus \mathcal{C}^*$. The concepts
        from $\mathcal{C}^*$ result in overall much higher accuracies, indicating that they better capture ideological polarization.}

        \label{fig:acc}
\end{figure}

Furthermore, we try whether it is possible to predict the DW-NOMINATE scores from the relative concept frequencies.
Specifically, we binarize the DW-NOMINATE scores by dividing them into the upper and lower half, thus resulting in a balanced dataset of more conservative and more liberal subreddits. 
We then train $\ell_2$-regularized 
logistic regression classifiers
using the relative frequencies of the concepts from $\mathcal{C}^*$ and $\mathcal{C} \setminus \mathcal{C}^*$ as features.
Since the dataset is small, 
we train 100 models on different random (label-stratified) splits of the subreddits into 50\% training and 50\% test. The models 
based on the concepts from $\mathcal{C}^*$ have substantially higher accuracies ($\mu = 0.657$, $\sigma = 0.122$)
than the models based on the concepts from $\mathcal{C} \setminus \mathcal{C}^*$ ($\mu = 0.491$, $\sigma = 0.109$), a difference that is again shown 
to be significant ($p < 0.01$) by a two-tailed $t$-test (Figure~\ref{fig:acc}). We interpret this as further evidence 
that the concepts in $\mathcal{C}^*$ (as opposed to the concepts in $\mathcal{C} \setminus \mathcal{C}^*$) capture ideological polarization.

\textbf{Qualitative analysis.} We analyze which concepts
are selected by SF-SGAE (Table \ref{tab:dynamics-examples}).
Many concepts in $\mathcal{C}^*$ are names of politicians (e.g., \textit{bush}, \textit{donald}) 
and designations of parties and political orientations (e.g., 
\textit{gop}, \textit{lefties}).
Furthermore, $\mathcal{C}^*$ contains concepts
related to contested political issues.
While many of these issues (e.g., \textit{gay marriage}, \textit{gun control}) 
have been shown to 
be characterized by polarized online discussions before
\citep{Lai.2015, Demszky.2019}, others (e.g., \textit{deregulation}, \textit{mainstream})
have been in the focus to a lesser degree, 
highlighting \slap4slip's
potential as an exploratory framework.

The design of our model also allows us to analyze 
in what way the concepts are polarized.
To do so,
we first examine the weight distribution of $\alpha^{(c_j)}$ for all $c_j \in \mathcal{C}^*$.
We notice that for the majority of concepts (roughly 80\%) $\alpha^{(c_j)} = 1$, 
i.e., the model uses only information about salience. 
Concepts with $\alpha^{(c_j)}= 1$ 
tend to be of immediate relevance for certain 
ideologies, leading to higher frequencies in relevant network regions.
For example, in communist subreddits,  discussion often revolves around fascism 
as the central opposing ideology, leading to higher
frequencies of \textit{fascist} than in other parts of the network
(Figure \ref{fig:fascist}).

\begin{table} [t!]\centering
\resizebox{\linewidth}{!}{%
\begin{tabular}{@{}llll@{}}
\toprule
Year & $\alpha^{(c_j)} = 0$ & $0 < \alpha^{(c_j)} < 1$  & $\alpha^{(c_j)} = 1$ \\
\midrule
\multirow{3}{*}{2013}& \textit{aca} (l/b) & \textit{deregulation} (l/b)  & \textit{gay marriage}  \\
& \textit{bush} (a/s) & \textit{fox news} (f/c) & \textit{gerrymandering}  \\
& \textit{tax} (c/h) & \textit{gun control}  (l/b) & \textit{surveillance} \\
\midrule
\multirow{3}{*}{2016}& \textit{julian} (l/b) & \textit{cuba} (a/s)  & \textit{collusion}  \\
& \textit{russian} (s/d) & \textit{gop} (s/d) & \textit{fake news}  \\
& \textit{trump voters} (c/h) & \textit{nationalism}  (l/b) & \textit{reagan} \\
\midrule
\multirow{3}{*}{2019} &  \textit{fact} (a/s)  & \textit{congress} (a/s)  & \textit{donald} \\
 & \textit{illegal} (a/s) &  \textit{white} (s/d) & \textit{fascist} \\
 & \textit{mainstream} (s/d)  &  \textit{women} (c/h) & \textit{lefties} \\
\bottomrule
\end{tabular}}
\caption{Example concepts with  $\alpha^{(c_j)}$
  values
of 1, 0, and in between.
For $\alpha^{(c_j)} < 1$, we also provide the moral foundation $m_k$ with maximum $\pi^{(c_j)}_k$. c/h: care/harm; f/c: fairness/cheating; l/b: loyalty/betrayal; a/s: authority/subversion; s/d: sanctity/degradation. \textit{aca} stands for Affordable Care Act (also known as Obamacare). \textit{julian} refers to Julian Assange.}  \label{tab:dynamics-examples}
\end{table}

\begin{table*} [t!]\centering
\resizebox{\linewidth}{!}{%
\begin{tabular}{@{}lllll@{}}
\toprule
 {} & \multicolumn{2}{c}{Small value of $|p_k(v_i, c_j)|$}  &  \multicolumn{2}{c}{Large value of $|p_k(v_i, c_j)|$}  \\
\cmidrule(lr){2-3}
\cmidrule(lr){4-5}
Concept $c_j$ & Subreddit $v_i$ & Example & Subreddit $v_i$ & Example \\
\midrule
\makecell[l]{ \textit{bush} \\ (2013, a/s)}& \texttt{Freethought}  & \makecell[{{p{6.4cm}}}]{\textit{This reminds me of what I read about the way the Bush administration worked religious quotes into military briefings.}}  &
\texttt{Anarchy101} & \makecell[{{p{6.4cm}}}]{\textit{What's stopping from murderers becoming presidents? Oh wait... US has Obama, previously had Bush.}} \\
\midrule
\makecell[l]{ \textit{trump voters} \\(2016, c/h)} & \texttt{Conservative}  & \makecell[{{p{6.4cm}}}]{\textit{Trump voters, and people on the right in general, believe this is a grand country with little institutional racism left.}} &
\texttt{socialism} & \makecell[{{p{6.4cm}}}]{\textit{Trump voters have a hate boner for the Clintons that they've maintained since their 92 campaign.}} \\
\midrule
\makecell[l]{  \textit{mainstream} \\(2019, s/d)} &  \texttt{Kamala} & \makecell[{{p{6.4cm}}}]{\textit{She's good at making progressive ideas sound like reasonable mainstream policies, which is the best of both worlds.
}  } &  \texttt{TheNewRight} & \makecell[{{p{6.4cm}}}]{\textit{I think mainstream media has infected your brain with such rot that it effects your emotions.}}  \\
\bottomrule
\end{tabular}}
\caption{Polarization in framing. The table provides contexts for three concepts with $\alpha^{(c_j)} = 0$, both for 
subreddits with weak framing ($|p_k(v_i, c_j)|$ small) and subreddits with strong framing
 ($|p_k(v_i, c_j)|$ large) in the relevant moral subspace.  c/h: care/harm; a/s: authority/subversion; s/d: sanctity/degradation.}  \label{tab:framing-examples}
\end{table*}

For concepts with $\alpha^{(c_j)} \neq 1$,
we can analyze which moral foundation has the largest $\pi^{(c_j)}_k$.
This moral foundation constitutes the basis for inter-subreddit differences in highlighting 
certain aspects of the concepts, which can be measured by $|p_k(v_i, c_j)|$, i.e., the absolute value 
of the projection of the concept embedding onto the $m_k$ subspace. 
For example, within the sanctity/degradation subspace (the subspace with maximal  $\pi^{(c_j)}_k$), 
many subreddits frame the concept \textit{mainstream} in neutral terms.
Right-wing subreddits, on the other hand, frame it
as something degenerate, particularly in the context of media (Figure~\ref{fig:mainstream}, Table~\ref{tab:framing-examples}),
reflecting appeals to discredit mainstream media reporting of political news \citep{Lee.2020}.

\begin{figure*}[t!]
        \centering
        \includegraphics[width=0.9\textwidth]{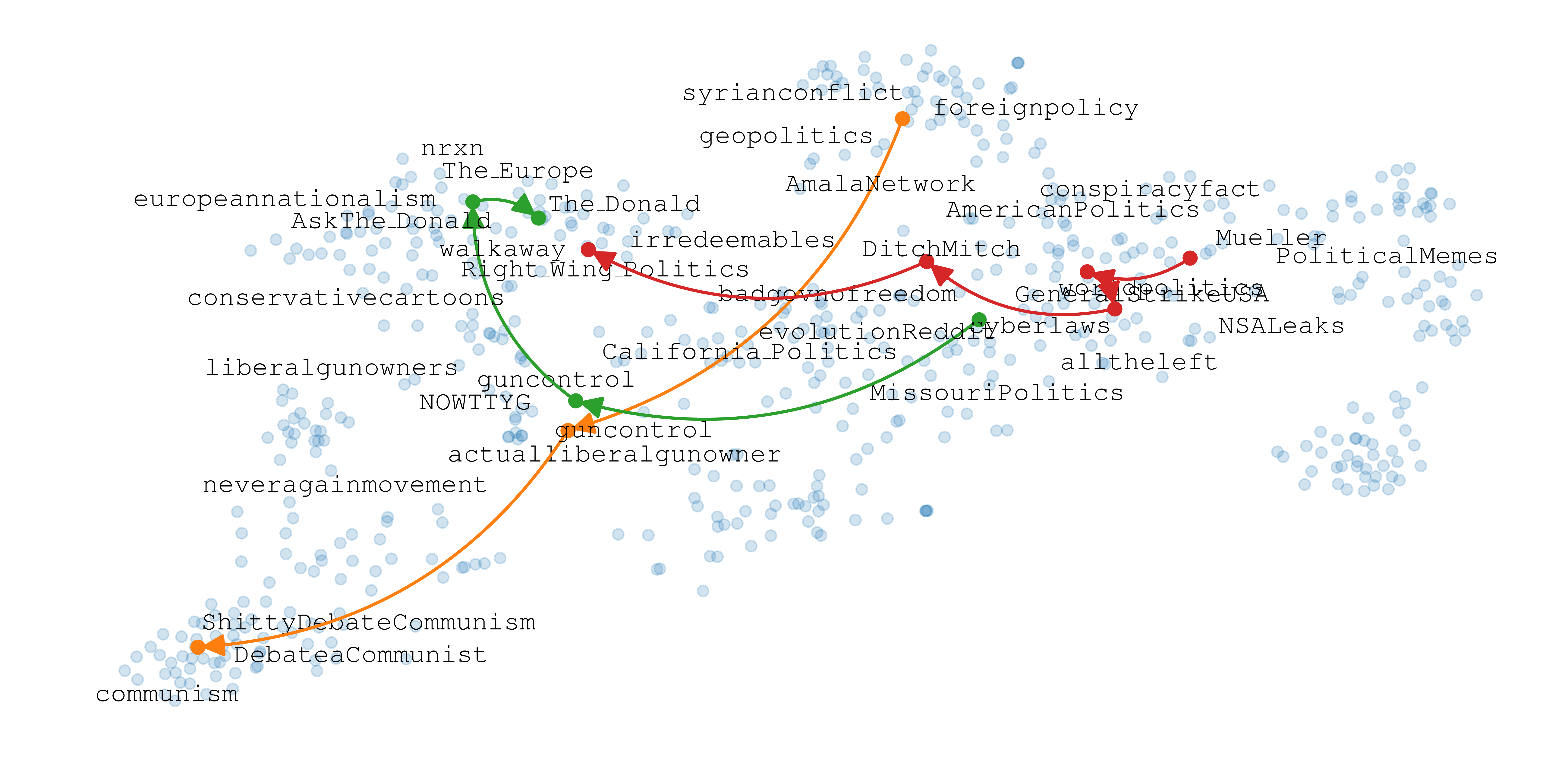}     
        \vspace{-0.3cm} 
        \caption[]{Example subreddits with a pronounced shift in their ideology over time. Orange: \texttt{Sino}, a subreddit
        originally devoted to geopolitics that moved to a more left-wing position; green and red: \texttt{FreeSpeech} and \texttt{POLITIC}, two originally moderate subreddits that moved to a more right-wing position.}        \label{fig:dynamics}
\end{figure*}

To get a more global picture of which moral subspaces
are most important for the polarized framing, we examine the learned values of $\pi^{(c_j)}_k$ 
(Section~\ref{sec:models}) for all concepts with $\alpha^{(c_j)} \neq 1$. The three moral foundations that most frequently have the highest $\pi^{(c_j)}_k$ value are
loyalty/betrayal
(30\%), sanctity/degradation (27\%), and authority/subversion (21\%), followed by care/harm (18\%) 
and fairness/cheating
(3\%). Interestingly, loyalty/betrayal,  sanctity/degradation, and
authority/subversion are the three moral foundations with
the greatest democrat-republican  differences \citep{Haidt.2007, Graham.2009},
indicating that the US two-party system is a central axis for the polarized framing of 
concepts on Reddit.

\textbf{Ideological dynamics.} The embeddings $\mathbf{Z}$ learned by our model are subreddit representations that combine linguistic 
information with network information. Here, we analyze 
what types of temporal ideological dynamics are captured by $\mathbf{Z}$.

We map the embeddings $\mathbf{Z}$ for all years into a common embedding space using orthogonal 
Procrustes \citep{Schonemann.1966, Hamilton.2016} and measure for each subreddit 
the cosine similarities between its embedding in the first year and its embeddings 
in all subsequent years. If the resulting time series
of cosine similarities is continuously decreasing, this indicates
a change in ideology.
To detect such shifts automatically, we compute for each subreddit Pearson's
$r$ between the time series of years and the time series of cosine similarities. Examining
the subreddits with the most extreme negative values of $r$, we observe that most of them 
experienced a pronounced shift in their ideological orientation (Figure \ref{fig:dynamics}). 
Specifically, the subreddits move from a relatively moderate to a more 
extreme position in ideology space, either right-wing (e.g., \texttt{FreeSpeech}, \texttt{POLITIC})
or left-wing (e.g., \texttt{Sino}). This pattern suggests that the subreddits have ideologically radicalized over time \citep{Grover.2019, Youngblood.2020}.

\section{Limitations}

The success of our method depends on how accurately 
polarization is reflected by the network, which means
that care must be taken during network selection (explicit networks) and construction (implicit networks).
For example, user overlap on Reddit can also be due to conflict between subreddits
\citep{Datta.2017, Kumar.2018, Datta.2019}. While
we do not find this to affect our results, it might
be a limitation if the degree of homophily in the network is too low.

This paper only applies \slap4slip to networks with communities as nodes 
and edges based on user overlap between the communities. However, the kind of clusteredness our method draws upon
has been shown to be a property of various types of social networks, including social networks with individual users as nodes such as 
Twitter \citep{Conover.2011, Himelboim.2013}.
We expect \slap4slip to be a suitable framework for finding polarized concepts in these cases, too.

\section{Conclusion}

We introduce \slap4slip (Sparse LAnguage Properties
for Social LInk Prediction), a novel framework 
for finding
linguistic features maximally 
informative about the structure of a social network, and show that it can be used to detect polarized concepts.
We model polarization along the dimensions of 
salience and framing.
While we only address polarized concepts in this paper, the general nature of the framework makes it possible 
to apply it in diverse scenarios involving linguistic data attached to social 
networks (e.g., to find the most pronounced topical differences in 
citation networks). 
We see 
our study as an exciting first step towards
bringing together 
computational social science research on online
polarization, NLP work 
on political language, and graph-based deep learning.

\section*{Ethical Considerations}

As part of our model, we use contextualized word embeddings to model the polarized framing of concepts. 
However, contextualized word embeddings are known to be biased \citep{Basta.2019, Zhao.2019, Bender.2021},
which bears the risk of impacting our results.
We see this as an important research question 
for future work.

The user base of Reddit has been shown to be disproportionately young 
and male compared to the general population of the US \citep{Shatz.2017}.
We acknowledge that this limits the generalizability of our results, 
and we try to be particularly careful when drawing conclusions in the paper.

\section*{Acknowledgements}

This work was funded by the European Research
Council (\#740516) and the Engineering and Physical
Sciences Research Council (EP/T023333/1).
The first author was supported by the German
Academic Scholarship Foundation and the Arts
and Humanities Research Council.
We thank the reviewers for
their helpful comments.
We are grateful to Jeremy Frimer for making the Moral Foundations Dictionary z-scores
available to us.

\bibliography{acl_latex}

\appendix

\section{Appendix}
\label{sec:appendix}

\subsection{Data Statistics} \label{app:stats}

Table~\ref{tab:data-stats} provides summary statistics of Reddit Politosphere \citep{Hofmann.2022}.
We compute average shortest path length as
\begin{equation*}
\mu_\pi = \sum_{i, j \in \mathcal{V}} \frac{\pi(i, j)}{|\mathcal{V}|(|\mathcal{V}| -1)},
\end{equation*}
where $\pi(i, j)$ is the shortest path from subreddit $i$ to subreddit $j$.
We compute density as
\begin{equation*}
\rho = \frac{2 |\mathcal{E}|}{|\mathcal{V}|(|\mathcal{V}| -1)}.
\end{equation*}
We compute modularity as
\begin{equation*}
Q = \frac{1}{2|\mathcal{E}|}  \sum_{i, j \in \mathcal{V}} \left( \mathbf{A}_{ij} -  \frac{d_i d_j}{2|\mathcal{E}|} \right) \delta(i, j),
\end{equation*}
where $\delta(i, j) = 1$ if $i$ and $j$ are in the same community, else $\delta(i, j) = 0$.
The maximum $Q$ values are indicative of the level of polarization
in the graph. $Q > 0.3$ for all years, which is a typical cut-off value
to determine polarized networks \citep{Garcia.2015}. Notice we use the standard definitions of the three measures \citep{Newman.2018}.

\begin{table} [t!]\centering
\resizebox{\linewidth}{!}{%
\begin{tabular}{@{}lrrrrrrr@{}}
\toprule
Year & $|\mathcal{D}|$ & $|\mathcal{V}|$ & $|\mathcal{E}|$ &  $\mu_d$  & $\mu_\pi$ & $\rho$ & $Q$\\
\midrule
2013 &6,306,458 & 108 & 324  & 6.00 & 3.08 & .056 & .560 \\
2014 &6,664,567 & 132 & 335  & 5.08 & 3.86 & .039 & .663 \\
2015 &9,230,022 & 168 & 493 & 5.87 & 3.87 & .035 & .672 \\
2016 &34,801,075 & 255 & 1,318  & 10.34 & 3.14 & .041 & .603  \\
2017 &38,278,685 & 295 & 1,572  & 10.66 & 3.14 & .036 & .585 \\
2018 & 40,222,627& 316 & 1,604 & 10.15 & 3.17 & .032 & .584 \\
2019 &46,590,000 & 412 & 2,536 & 12.31 & 3.20 & .030 & .603\\

\bottomrule
\end{tabular}}
\caption{Dataset statistics. $|\mathcal{D}|$: number of comments; $|\mathcal{V}|$: number of nodes (subreddits); $|\mathcal{E}|$: number of 
edges; $\mu_d$: average node degree; $\mu_\pi$: average shortest path length; $\rho$: density; $Q$: maximum modularity.}  \label{tab:data-stats}
\end{table}

\begin{table*} [t!]\centering
\resizebox{0.92\linewidth}{!}{%
\begin{tabular}{@{}lrrrrrrrrrrrrrrrr@{}}
\toprule
{} & \multicolumn{8}{c}{Fairness/cheating} & \multicolumn{8}{c}{ Sanctity/degradation } \\
\cmidrule(lr){2-9}
\cmidrule(l){10-17}

Set & 2013 & 2014 & 2015 & 2016 & 2017 & 2018 & 2019 & $\mu\pm\sigma$ & 
2013 & 2014 & 2015 & 2016 & 2017 & 2018 & 2019 & $\mu\pm\sigma$\\
\midrule
 $\mathcal{T}_k(c_j)$ & \best{.074} & \best{.074} & \best{.074} & \best{.075} & \best{.076} & \best{.076} & \best{.076} & .075$\pm$.001 & \best{.067} & \best{.068} & \best{.067} & \best{.070} & \best{.069} & \best{.068} & \best{.067} & .068$\pm$.001\\ 
$\mathcal{B}_k(c_j)$ & .068 & .068 & .068 & .070 & .071 & .070 & .070 & .069$\pm$.001 & .065 & .064 & .064 & .067 & .066 & .065 & .064 & .065$\pm$.001\\  
\bottomrule
\end{tabular}}
\caption{Comparison of average $p_k(v_i, c_j)$ values for $\mathcal{T}_k(c_j)$ (large proportion of moral context words) and $\mathcal{B}_k(c_j)$ (small proportion of moral context words). The table shows the values for fairness/cheating and sanctity/degradation, but the trend is consistent across all moral foundations. Higher value per column in gray.}  
\label{tab:proj}
\end{table*}

\subsection{Details on Moral Subspaces} \label{app:subspaces}

For $\mathbf{e}(v_i, c_j)$, we extract the mean-pooled embedding
if the concept is split into multiple WordPiece tokens and sample a maximum of 100 occurrences per subreddit and concept. 
For $\mathbf{e}(m_k)$, we sample 1,000 occurrences per word.

It is important to notice that
$p_k(v_i, c_j)$ is impacted by two different factors.
On the one hand, $p_k(v_i, c_j)$ captures the association of concepts with moral foundations due to 
\textit{intrinsic} lexical-semantic properties, which can be seen by examining the variation of 
$p_k(v_i, c_j)$ across different concepts. Thus, computing
$\frac{1}{|\mathcal{V}|} \sum_{v_i \in \mathcal{V}} p_k(v_i, c_j)$ for all concepts and moral foundations (i.e., 
the average value of $p_k(v_i, c_j)$ across subreddits), we find that the lexical semantics of 
concepts with the highest values are directly related to the moral foundations (e.g., \textit{patriot}
and \textit{revolution} for loyalty/betrayal). 

On the other hand,
$p_k(v_i, c_j)$ also captures the association of concepts with moral foundations that is due to 
\textit{extrinsic} cooccurrence patterns caused by ideological framing, which can be seen by examining the variation of $p_k(v_i, c_j)$ across different
contexts and subreddits (i.e., sets of contexts). To check this empirically, we 
use the 20 highest-ranked words per
moral foundation from the Moral Foundations Dictionary \citep{Frimer.2017} and compute for each subreddit $v_i$, concept $c_j$, and moral foundation $m_k$ the proportion of occurrences 
in which at least one $m_k$ word is found in a context window of 10 words around $c_j$, which is similar to traditional
ways of measuring ideological framing (e.g., \citealp{Fulgoni.2016}). 
We then create for each concept $c_j$ and moral foundation $m_k$ (i) a set $\mathcal{T}_k(c_j)$
containing the $d$ subreddits with the largest proportion of moral context words
and (ii) a set $\mathcal{B}_k(c_j)$ containing the $d$ subreddits with the smallest proportion of moral context words. We set $d=5$, but results are robust with respect to the exact selection of $d$. 
Comparing the average value of $p_k(v_i, c_j)$ of subreddits in $\mathcal{T}_k(c_j)$
and $\mathcal{B}_k(c_j)$ for all concepts,
we find it to be consistently higher for $\mathcal{T}_k(c_j)$
than for $\mathcal{B}_k(c_j)$ (Table \ref{tab:proj}). The fact that this result holds for all years and moral foundations
suggests that the extent to which the concepts cooccur with 
certain moral frames is indeed captured by the projections of
contextualized embeddings into the moral subspaces.
Crucially, while $p_k(v_i, c_j)$ in principle captures both types of factors, only the 
extrinsically-driven variation due to ideological framing is expected to 
be valuable for predicting the social network structure.

\subsection{Hyperparameters}\label{app:hyperparams}

The input layer of the model has 1,000 dimensions (which are sparsified during training),
the first hidden layer 100 dimensions, and the second hidden layer 
10 dimensions. We perform grid search for the number of epochs $e \in \{1, \dots ,1000\}$, the learning rate $r \in \{ \num{1e-4}, \num{3e-4} , \num{1e-3}, \num{3e-3} \}$, and
the regularization constant $\lambda \in \{ \num{1e-4}, \num{3e-4} , \num{1e-3}, \num{3e-3} \}$. 

All experiments are performed on a GeForce GTX 1080 Ti GPU (11GB).
The total number of trainable parameters
is 107,110 for SF-SGAE, SF-SLAE, and SF-GAE, 101,110 for S-SGAE, and 106,110 for F-SGAE.

\subsection{Dev Performance}\label{app:dev}

Table \ref{tab:auc-dev} provides the dev performance 
for all models considered in Section \ref{sec:experiments}
of the paper.

\begin{table} [t!]\centering
\resizebox{\linewidth}{!}{%
\begin{tabular}{@{}lrrrrrrrr@{}}
\toprule
Model & 2013 & 2014 & 2015 & 2016 & 2017 & 2018 & 2019 & $\mu\pm\sigma$\\
\midrule
SF-SGAE & \best{.857} & \best{.893} & \best{.911} & .921 & .923 & .913 & .921 & .906$\pm$.022 \\
\midrule 
S-SGAE & .833 & .868 & .872 & .864 & .883 & .865 & .904 & .870$\pm$.020 \\ 
F-SGAE & .832 & .880 & .863 & .861 & .884 & .868 & .894 & .869$\pm$.019\\ 
SF-SLAE & .712 & .812 & .772 & .771 & .778 & .729 & .748 & .760$\pm$.031 \\ 
\midrule
SF-GAE & .852 & .887 & .910 & \best{.935} & \best{.939} & \best{.926} & \best{.943} & .913$\pm$.031\\ 
\bottomrule
\end{tabular}}
\caption{Dev performance (AUC). SF-SGAE outperforms S-SGAE, F-SGAE, and SF-SLAE. 
It performs similarly to or better than SF-GAE despite using only a fraction of concepts. 
Best score per column in gray.}  \label{tab:auc-dev}
\end{table}

\subsection{Sparsity Threshold} \label{app:sparsity}

\begin{figure*}[t!]
        \centering
        \includegraphics[width=0.9\textwidth]{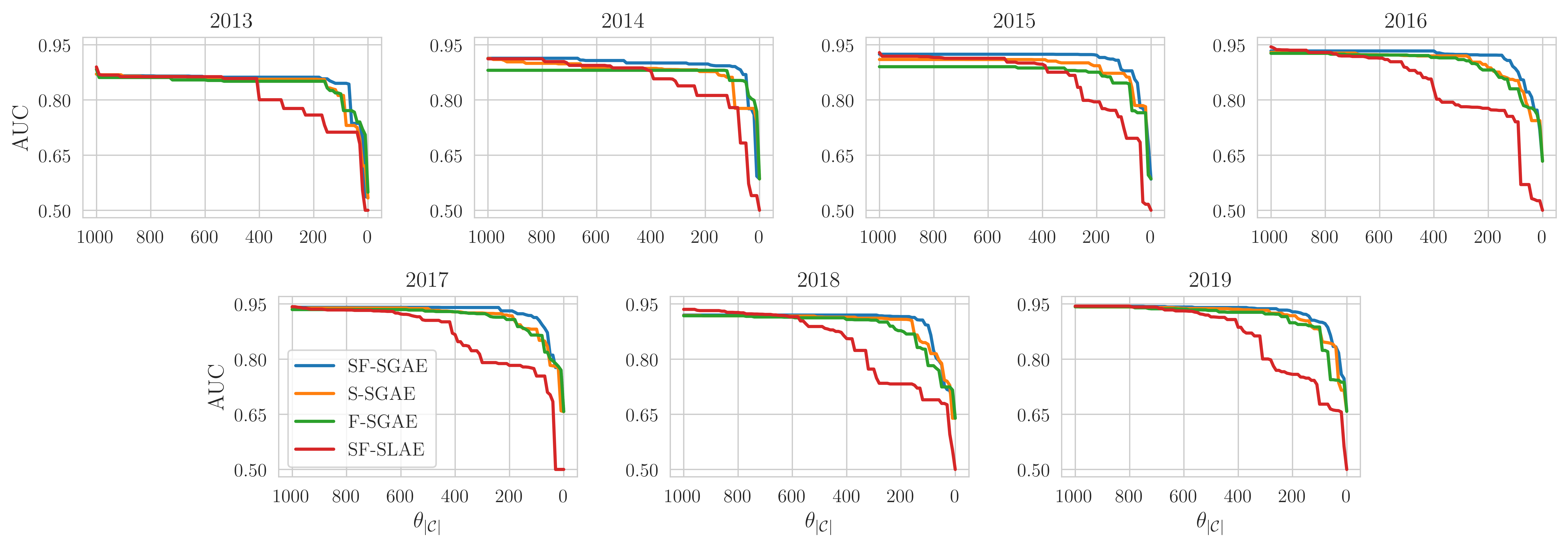}      
        \caption[]{Impact of  sparsity threshold
          $\theta_{|\mathcal{C}|}$ on performance (AUC). 
          SF-SGAE performs better than any other model in the sparse regime ($\theta_{|\mathcal{C}|} \leq 200$), 
        showing that it better captures polarization.}
        \label{fig:threshold_all}
\end{figure*}

Figure \ref{fig:threshold_all} presents the results of the experiment varying the sparsity threshold 
described in Section \ref{sec:experiments} of the paper for all years.

\end{document}

%% file: tutorial_1.tex
\begin{tikzpicture}
[xscale=0.9,
 thick,
 vertex1/.style={circle,fill=blue!90,
inner sep=0pt,minimum size=5mm},
 vertex10/.style={circle,fill=red!90,
inner sep=0pt,minimum size=5mm}]

\node[vertex1](s_1){};
\node[vertex1](s_2)[left=1cm of s_1] {};
\node[vertex1](s_3) [below=1cm of s_1] {};
\node[vertex1](s_4)[below=1cm of s_2]  {};

\node[vertex10](s_5)[right=0.3cm of s_1]  {};
\node[vertex10](s_6) [right=1cm of s_5] {};
\node[vertex10](s_7) [below=1cm of s_5]  {};
\node[vertex10](s_8) [below=1cm of s_6]  {};

\draw[](s_1) to (s_2);
\draw[](s_1) to (s_3);
\draw[](s_1) to (s_4);
\draw[](s_2) to (s_3);
\draw[](s_2) to (s_4);
\draw[](s_3) to (s_4);

\draw[](s_5) to (s_6);
\draw[](s_5) to (s_7);
\draw[](s_5) to (s_8);
\draw[](s_6) to (s_7);
\draw[](s_6) to (s_8);
\draw[](s_7) to (s_8);

\end{tikzpicture}

%% file: tutorial_2.tex
\begin{tikzpicture}
[xscale=0.9,
 thick,
 vertex1/.style={circle,fill=blue!90,
inner sep=0pt,minimum size=5mm},
 vertex10/.style={circle,fill=red!90,
inner sep=0pt,minimum size=5mm}]

\node[vertex10](s_1){};
\node[vertex1](s_2)[left=1cm of s_1] {};
\node[vertex10](s_3) [below=1cm of s_1] {};
\node[vertex1](s_4)[below=1cm of s_2]  {};

\node[vertex10](s_5)[right=0.3cm of s_1]  {};
\node[vertex1](s_6) [right=1cm of s_5] {};
\node[vertex1](s_7) [below=1cm of s_5]  {};
\node[vertex10](s_8) [below=1cm of s_6]  {};

\draw[](s_1) to (s_2);
\draw[](s_1) to (s_3);
\draw[](s_1) to (s_4);
\draw[](s_2) to (s_3);
\draw[](s_2) to (s_4);
\draw[](s_3) to (s_4);

\draw[](s_5) to (s_6);
\draw[](s_5) to (s_7);
\draw[](s_5) to (s_8);
\draw[](s_6) to (s_7);
\draw[](s_6) to (s_8);
\draw[](s_7) to (s_8);

\end{tikzpicture}